\documentclass[a4paper,fleqn]{cas-dc}
\usepackage[authoryear]{natbib}
\usepackage[utf8]{inputenc}
\usepackage{times}
\usepackage{epsfig}
\usepackage{graphicx}
\usepackage{amsmath}
\usepackage{amssymb}
\usepackage{color}
\usepackage{multirow}
\usepackage{algorithmic}
\usepackage{algorithm}
\usepackage{bbding} 
\usepackage{array}
\usepackage{textcomp}
\usepackage{stfloats}
\usepackage{url}
\usepackage{verbatim}
\usepackage{graphicx}
\usepackage{tablefootnote}
\usepackage{ulem} 
\usepackage{mathrsfs}
\usepackage[switch]{lineno}

\usepackage{anyfontsize}

\makeatletter
\renewcommand{\fnum@figure}{Fig. \thefigure.\@gobble}
\makeatother

\def\tsc#1{\csdef{#1}{\textsc{\lowercase{#1}}\xspace}}
\tsc{WGM}
\tsc{QE}
\makeatletter\def\Hy@Warning#1{}\makeatother
\begin{document}
\let\WriteBookmarks\relax
\def\floatpagepagefraction{1}
\def\textpagefraction{.001}

\shorttitle{3D Learnable Supertoken Transformer for LiDAR Point Cloud Scene Segmentation}    

\shortauthors{D. Lu et al.}

\title[mode = title]{3D Learnable Supertoken Transformer for LiDAR Point Cloud Scene Segmentation}

\author[1]{Dening Lu}[orcid=0000-0003-0316-0299]

\author[2]{Jun Zhou}

\author[1]{Kyle (Yilin) Gao}


\author[1]{Linlin Xu}[orcid=0000-0002-3488-5199]
\cormark[1]
\cortext[1]{Corresponding authors. 
l44xu@uwaterloo.ca (Linlin Xu), junli@uwaterloo.ca (Jonathan Li)}

\author[3]{Jonathan Li}
\cormark[1]

\begin{abstract}
3D Transformers have achieved great success in point cloud understanding and representation. However, there is still considerable scope for further development in effective and efficient Transformers for large-scale LiDAR point cloud scene segmentation. This paper proposes a novel 3D Transformer framework, named \textbf{3D} \textbf{L}earnable \textbf{S}upertoken \textbf{T}ransformer (\textbf{3DLST}). The key contributions are summarized as follows. Firstly, we introduce the first Dynamic Supertoken Optimization (DSO) block for efficient token clustering and aggregating, where the learnable supertoken definition avoids the time-consuming pre-processing of traditional superpoint generation. Since the learnable supertokens can be dynamically optimized by multi-level deep features during network learning, they are tailored to the semantic homogeneity-aware token clustering. Secondly, an efficient Cross-Attention-guided Upsampling (CAU) block is proposed for token reconstruction from optimized supertokens. Thirdly, the 3DLST is equipped with a novel W-net architecture instead of the common U-net design, which is more suitable for Transformer-based feature learning. The SOTA performance on three challenging LiDAR datasets (airborne MultiSpectral LiDAR (MS-LiDAR) (89.3$\%$ of the average $F_{1}$ score), DALES (80.2$\%$ of mIoU), and Toronto-3D dataset (80.4$\%$ of mIoU)) demonstrate the superiority of 3DLST and its strong adaptability to various LiDAR point cloud data (airborne MS-LiDAR, aerial LiDAR, and vehicle-mounted LiDAR data).  Furthermore, 3DLST also achieves satisfactory results in terms of algorithm efficiency, which is up to 5$\times$ faster than previous best-performing methods.
\end{abstract}

\maketitle
\section{Introduction}
\label{sec:introduction}
LiDAR (Light Detection and Ranging) point cloud scene segmentation is an important technique in remote sensing with wide-ranging implications. As our world increasingly relies on remote sensing technologies for applications in urban planning \citep{stilla2023change, han2024whu}, environmental monitoring \citep{han2023survey, li2024efficient}, disaster management \citep{schuegraf2024planes4lod2, wang2020extraction}, and more \citep{ chen2023adaptive}, the need for accurate and efficient segmentation of LiDAR point clouds becomes critical. By segmenting these vast and complex datasets into meaningful components such as buildings, trees, roads, and ground, we are able to extract valuable information for various purposes, including 3D modeling \citep{li2022point2roof}, land cover classification \citep{wang2023imbalance}, and infrastructure analysis \citep{lin2022semantic}. Moreover, precise segmentation enables automated processing and analysis, facilitating rapid decision-making and resource allocation in diverse fields. 

The Transformer technique has achieved great success in various research fields such as Natural Language Processing (NLP) \citep{NawrotCLP23}, image processing \citep{DosovitskiyB0WZ21, liu2021swin}, and point cloud representation and understanding \citep{zhao2021point, guo2021pct, lai2022stratified}. Initially pioneered in NLP tasks, Transformers have demonstrated remarkable capabilities in capturing long-range dependencies and contextual information, which are crucial for understanding complex spatial relationships within point cloud data. In point cloud segmentation, traditional methods usually rely on Convolutional Neural Networks (CNNs), which struggle to effectively capture global context due to the locality inductive bias. However, 3D Transformer architectures, such as the Point Transformer and Point Cloud Transformer \citep{zhao2021point, guo2021pct}, have shown promising results by directly operating on unordered point sets, enabling effective modeling of interactions between points regardless of their spatial arrangement. This approach enhances the model's ability to discern intricate structures within point clouds, leading to more accurate segmentation results. Further, to address the issue of the high computational footprint and memory consumption of 3D Transformers, many efficient Transformer architectures \citep{hui2021pyramid, zhang2022patchformer, park2022fast, sun2023superpoint, robert2023efficient, wang2023dsvt, liu2023flatformer} have been proposed. Most recently, integrating superpoints \citep{landrieu2018large} with 3D Transformer techniques \citep{sun2023superpoint, robert2023efficient} has achieved tremendous progress in terms of both efficiency and accuracy improvement of point cloud segmentation. The superpoint clustering strategy reduces the number of input tokens to the Transformer network significantly and provides relatively accurate over-segmentation results, which facilitates the subsequent feature learning and understanding.

Despite the great improvement achieved by superpoint Transformers, the limited initial point features involved in superpoint construction and its time-consuming computations (Please refer to Section \ref{sec:relatedwork} for detailed analysis) tend to hinder further improvement of model performance. To address this issue, we proposed a novel supertoken Transformer framework for point cloud segmentation, named 3DLST, which is the first work to introduce learnable supertoken and dynamic optimization in the 3D Transformer paradigm. Compared with the aforementioned static superpoint clustering method, the proposed DSO block is able to optimize the supertokens dynamically based on multi-level deep features. Therefore, the learnable supertokens are beneficial to the semantic homogeneity-aware token clustering process, which is more flexible and efficient. Moreover, we proposed a novel W-net architecture tailored to 3D Transformer-based point cloud segmentation, which outperforms the traditional U-net design in all our experiments.

In summary, the main contributions of our work are as follows:
\begin{itemize}
    \item We propose the first Dynamic Supertoken Optimization (DSO) block in the 3D Transformer paradigm for efficient token clustering and aggregating. The proposed learnable supertokens serve the semantic homogeneity-aware token clustering process directly, outperforming traditional superpoint methods. 
    \item An efficient Cross-Attention-guided Upsampling (CAU) block is proposed for token reconstruction. It fully exploits the long-range context dependency modeling capability of the Transformer, achieving the upsampling process by rich semantic similarity information among tokens. 
    \item By integrating the aforementioned blocks, we present a 3D Learnable Supertoken Transformer (3DLST) network, equipped with a novel W-net architecture tailored to 3D Transformer-based point cloud segmentation. Extensive experiments on various LiDAR point cloud datasets demonstrate the SOTA performance of 3DLST in terms of both accuracy and efficiency, as well as its strong adaptability to various LiDAR point cloud data (airborne MS-LiDAR, aerial LiDAR, and vehicle-mounted LiDAR data).
    
\end{itemize}

The remainder of our paper is organized as follows. Section \ref{sec:relatedwork} reviews existing 3D Transformer methods for point cloud segmentation and summarizes the limitations. Section \ref{sec:method} shows the details of 3DLST. Section \ref{sec:Experiments} presents and discusses the experimental results. Section \ref{sec:conclusion} concludes the paper.

\section{Related Work}
\label{sec:relatedwork}
In this section, we review the development and applications of 3D Transformers in point cloud segmentation and summarize the challenges they confront.

\subsection{Point Transformers}
Transformers have made significant contributions in the field of point cloud representation. Point Cloud Transformer (PCT) \citep{guo2021pct} and Point Transformer (PT) \citep{zhao2021point}, as pioneers of the 3D Transformer technique, have proven the great potential and high adaptability of Transformers in the field of point cloud processing. After that, a series of variants of point Transformer architectures \citep{qiu2021geometric,lu20223dctn,lai2022stratified,lu20243dgtn} were proposed to improve the model performance. GBNet proposed in \cite{qiu2021geometric} introduced a channel-wise self-attention mechanism to the point Transformer network. It designed a Channel-wise Affinity Attention (CAA) module for better feature representation, generating the similarity matrix in the channel space. A Channel Affinity Estimator (CAE) block was then proposed to calculate an affinity matrix, sharpening the attention weights and avoiding aggregating similar/redundant information. 3DCTN \citep{lu20223dctn} proposed to combine the graph convolution \citep{wang2019dynamic} and Transformer blocks for local feature extraction and global feature learning respectively, providing a promising solution for both accuracy and efficiency improvement of point Transformers. Besides, 3DCTN also conducted a detailed investigation on self-attention operators in 3D Transformers, including scalar attention and different forms of vector attention. Inspired by Swin Transformer \citep{liu2021swin}, Stratified Transformer (ST) \citep{lai2022stratified} split the point cloud into a group of non-overlapping cubic windows via 3D voxelization, followed by performing the self-attention mechanism in each window. To further improve the connection among independent windows, ST generated the dense local \textit{key} points and sparse distant \textit{key} points for each \textit{query} point, enlarging the receptive field of \textit{query} points for long-range context dependency modeling.

To improve the algorithm efficiency and reduce memory consumption, a series of lightweight point Transformers \citep{wu2021centroid, hui2021pyramid, cheng2023transrvnet, sun2023superpoint, robert2023efficient} were proposed. Centroid Transformer \citep{wu2021centroid} constructed $M$ centroids from $N$ input points, which are taken as $Query$ points of the Transformer blocks. The centroid construction is achieved by soft K-means optimization. As such, the computational complexity is reduced from $\mathcal{O}(N^{2}D)$ to $\mathcal{O}(NMD)$, where $M \ll N$. PPT-Net \citep{hui2021pyramid} proposed a hierarchical encoder-decoder Transformer network to improve model efficiency by gradually reducing the number of points. It also combined graph convolution-based local feature embedding and Transformer-based global feature learning, which allows PPT-Net to achieve a good balance between model efficiency and accuracy. TransRVNet \citep{cheng2023transrvnet} is proposed for efficient 3D LiDAR point cloud semantic segmentation, which is designed as a projection-based CNN-Transformer architecture to infer point-wise semantics. A Multi Residual Channel Interaction Attention Module (MRCIAM) is introduced to ensure the efficient process of channel-wise feature learning. Most recently, there are several superpoint Transformers \citep{sun2023superpoint, robert2023efficient} proposed for efficient token clustering and aggregating, resulting in a great reduction of input tokens to the Transformer network. They have seen tremendous progress in model efficiency improvement while achieving satisfactory point cloud segmentation results. The detailed description and discussion are shown in the following section (Sec .\ref{subsec:superpoint}).

\begin{figure*}[htbp]
  \centering
  \includegraphics[width=\linewidth]{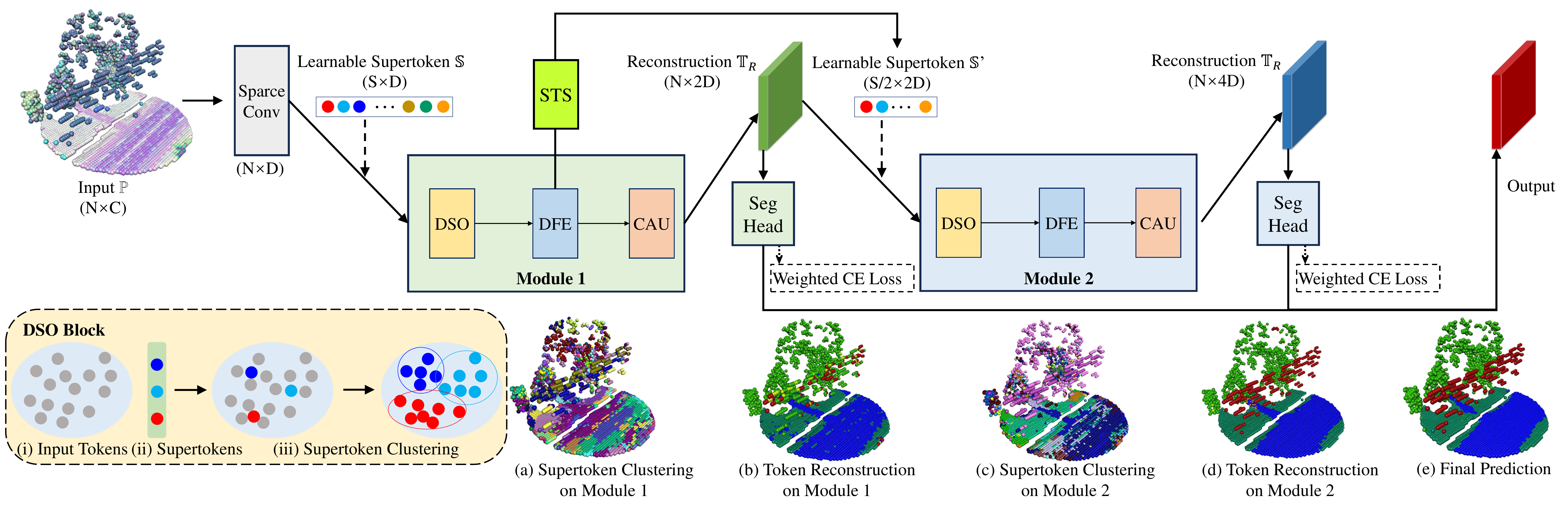}
  \caption{Architecture of 3DLST for point cloud segmentation. It has two main modules for token clustering and deep feature extraction, constituting the novel W-net structure. Specifically, DSO represents Dynamic Supertoken Optimization. DFE represents Deep Feature Enhancement. CAU represents Cross-attention-guided Upsampling. STS represents SuperToken Sparsification. The supertoken clustering and token reconstruction results on each module show the dynamic optimization process of the network. Besides, we also provide a brief illustration of the DSO block.
  \label{fig:overview}}
\end{figure*}

\subsection{Superpoint Transformers}
\label{subsec:superpoint}
The superpoint clustering strategy \citep{landrieu2018large}, as a strong algorithm for point cloud over-segmentation, has been wildly used in efficient algorithm designing for large-scale point cloud data processing. It takes the raw point clouds with limited initial features (geometric/color/intensity) as input, followed by partitioning the point clouds into a series of geometrically homogeneous groups. This process is formulated as a \textit{generalized minimal partition problem} in \cite{landrieu2018large}, where the ${\ell}_{0}$ pursuit algorithm \citep{landrieu2017cut} could find an approximate solution quickly. Since the superpoint generation only requires the initial features of input points, it is usually used as a pre-processing step to guide the following clustering process in the network, which is called the \textit{static superpoint} strategy in our paper. Recently, there have been several efficient Transformer methods \citep{sun2023superpoint, robert2023efficient} that were designed based on the combination of superpoint and Transformer networks. SPFormer \citep{sun2023superpoint} designed a superpoint pooling layer based on pre-computed superpoints to guide the point cloud downsampling, which significantly reduces the computational overhead of the following feature learning process. SPT \citep{robert2023efficient} applied the static superpoint strategy to the input point clouds, generating a hierarchical superpoint structure by adjusting the granularity of superpoint clustering. It allows the model to exploit the context at different scales for semantic feature understanding, resulting in large receptive fields and high efficiency. 

Despite the great success of superpoint Transformers in point cloud processing, they still have some drawbacks that limit the development of model efficiency and accuracy improvement. Firstly, as a pre-processing step, superpoint generation is very time-consuming, usually taking up more than 99$\%$ of the inference time. 
Secondly, only taking the initial point features as input, it is challenging for the static superpoint strategy to serve the semantic homogeneity clustering of deep features well. This tends to hinder the performance improvement. To address these drawbacks, we proposed a novel Transformer framework for point cloud segmentation, named 3D Learnable Supertoken Transformer (3DLST). Instead of generating superpoints according to the limited initial features, we define a series of learnable supertokens at the beginning of the network training. They could achieve deep feature clustering adaptively at different levels and be optimized dynamically during network learning. Without time-consuming superpoint construction, the pre-processing time of the method would also be reduced significantly. Extensive comparison experiments on various LiDAR datasets demonstrate that our method outperforms the previous SOTA superpoint Transformers in both model efficiency and accuracy.

\begin{figure*}[htbp]
  \centering
  \includegraphics[width=0.9\linewidth]{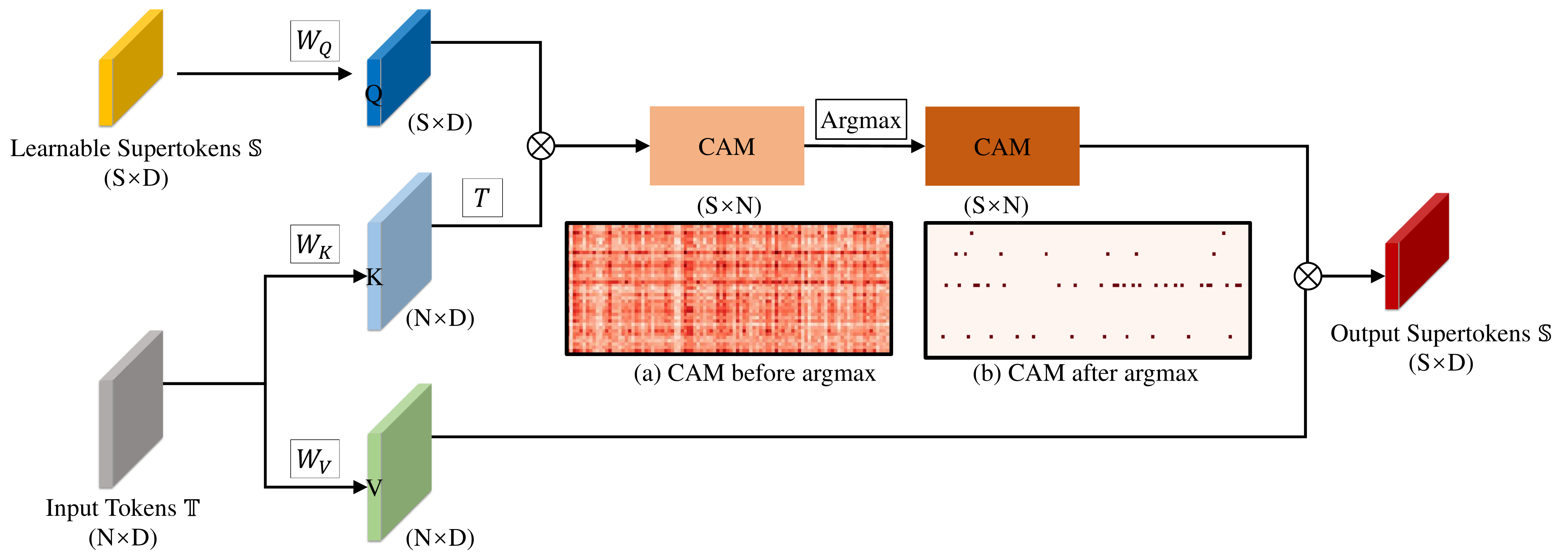}
  \caption{Illustration of the cross-attention calculation, where visualization results of CAM before/after argmax are shown clearly.
  \label{fig:cam}}
\end{figure*}

\section{3D Learnable Supertoken Transformer}
\label{sec:method}
We detail the 3DLST framework in this section. Firstly, we introduce the overall design of 3DLST. Secondly, we present the key components in 3DLST: Dynamic Supertoken Optimization (DSO),  Deep Feature Enhancement (DFE), Cross-attention-guided Upsampling (CAU), and SuperToken Sparsification (STS) blocks.

\subsection{Overview}
The overall framework of 3DLST is shown in Fig. \ref{fig:overview}. It takes the raw LiDAR point cloud $\mathbb{P}=\left\{ p_{i} \right\}_{i \in N} \in R^{N \times C}$ as input, where $C$ represents the feature dimension of input points. In 3DLST, we first apply a powerful and efficient feature extractor achieved by Point-Voxel CNN (PVCNN) \citep{liu2019point} for token embedding. The embedded tokens are denoted as $\mathbb{T}=\left\{ t_{i} \right\}_{i \in N} \in R^{N \times D}$, where $N \times D$ represents the feature dimension of embedded features. Secondly, two cascading modules for deep feature learning are proposed, constituting the novel W-net structure for point cloud representation. 

Specifically, for Module 1, we first propose a novel supertoken generation and optimization strategy (i.e., DSO) for token clustering and aggregating. It starts from randomly initialized learnable embeddings, defined as initial supertoken. Then a hard-assignment cross-attention mechanism is applied for dynamic token clustering based on the initial supertoken. After that, an updated supertoken set could be obtained by an average pooling operated on each cluster, followed by further enhancement via the DFE block. We then propose a novel and efficient upsampling block (i.e., CAU) for token reconstruction. Given the enhanced supertoken set, an efficient STS block is proposed before Module 2, to further reduce supertokens for efficiency improvement. As shown in Fig. \ref{fig:overview}, Module 2 has a similar structure to Module 1, and outputs reconstructed tokens with higher-dimension features than Module 1.

After using the two modules in a hierarchical manner, we could obtain two sets of reconstructed tokens with different feature dimensions. Finally, we apply an MLP-based segmentation head to each token set and adopt a multiloss strategy for network training. During testing, we compute the prediction probabilities addition of two token sets as the final results.

\subsection{Dynamic Supertoken Optimization (DSO)}
\label{subsec:DSO}
Taking the embedded tokens from the feature extractor as input, the DSO block defines a learnable supertoken set as clustering centers, denoted as $\mathbb{S}=\left\{ s_{i} \right\}_{i \in S} \in R^{S \times D}$. $\mathbb{S}$ is randomly initialized and gradually optimized during network training. Compared with the traditional superpoint constructing approaches which are usually based on limited initial attributes of the input points, DSO serves the semantic homogeneity-aware token clustering process directly, which is more flexible and efficient.

As shown in Fig. \ref{fig:cam}, given the embedded tokens $\mathbb{T}$ and the supertoken set $\mathbb{S}$, a cross-attention is performed to explore the semantic similarity between $\mathbb{T}$ and $\mathbb{S}$. Firstly, The $Query$, $Key$, and $Value$ matrices are computed as below:
\begin{equation}
\label{eq:qkv}
\begin{aligned}
Q =  \mathbb{S} \times W_{Q},\\
K =  \mathbb{T} \times W_{K} ,\\
V =  \mathbb{T} \times W_{V} , \\
\end{aligned}
\end{equation}
where $Q, K, V$ represents $Query$, $Key$, and $Value$ matrices, and $W_{Q}, W_{K}, W_{V}$ are learnable weight matrices. After that, a Cross-Attention Map (CAM) is calculated to reflect the feature similarity between $Q$ and $K$:
\begin{equation}
\label{eq:cam}
CAM = \mathop{\arg\max}\limits_{N}(\frac{Q\times K^{T}}{\sqrt{D}}).
\end{equation}
From Eq. \ref{eq:cam}, the $argmax$ operation performs a hard assignment from the $K$ set to the $Q$ set, which is inspired by the K-means clustering strategy \citep{yu2022k}. Ablation experiments (Section \ref{subsec:ablation}) in our paper show the $argmax$ operation achieves better results than $softmax$ in our framework. Finally, the supertoken set $\mathbb{S}$ could be improved by averaging tokens in the same cluster:
\begin{equation}
\label{eq:improve_s}
\mathbb{S} = \frac{CAM \times V}{\sum_{j=1}^{N} CAM_{ij}},
\end{equation}
where $CAM_{ij}$ represent the $i$-th row and $j$-th column element in CAM. From Eq. \ref{eq:improve_s}, the updated $\mathbb{S}$ have the same size as that in Eq. \ref{eq:qkv}, i.e., $S \times D$.

\begin{figure*}[htbp]
  \centering
  \includegraphics[width=0.9\linewidth]{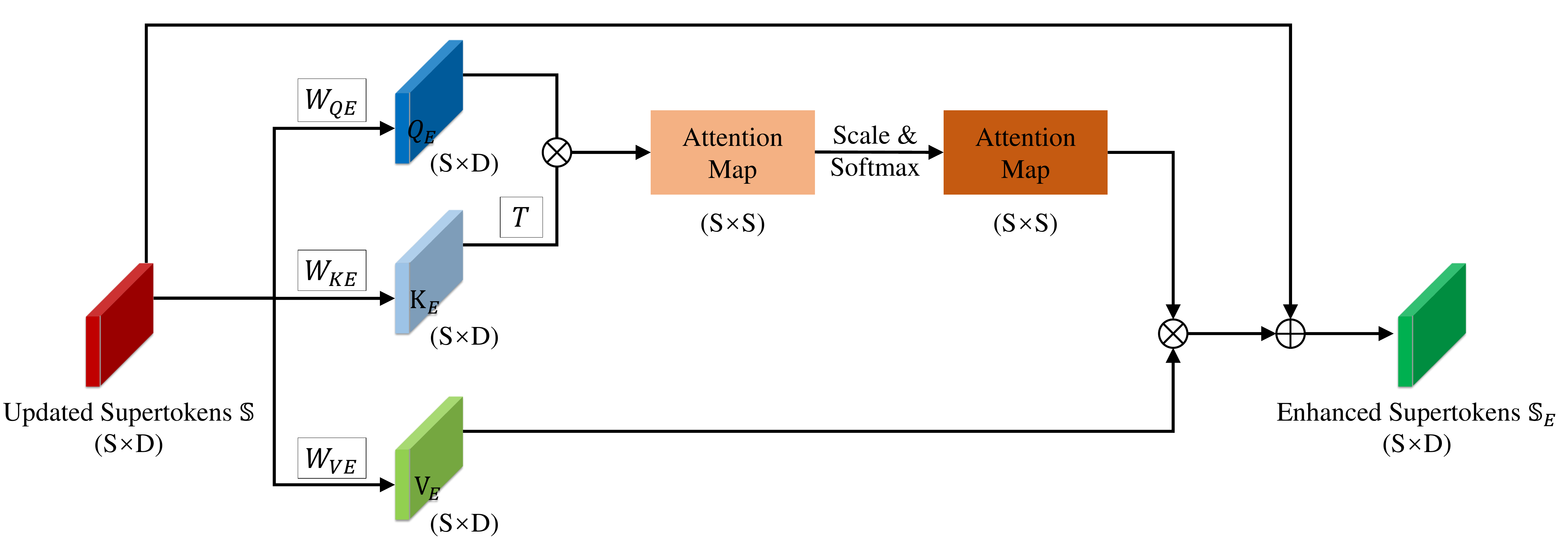}
  \caption{Illustration of the DFE block.
  \label{fig:dfe}}
\end{figure*}

\begin{figure}[b]
  \centering
  \includegraphics[width=\linewidth]{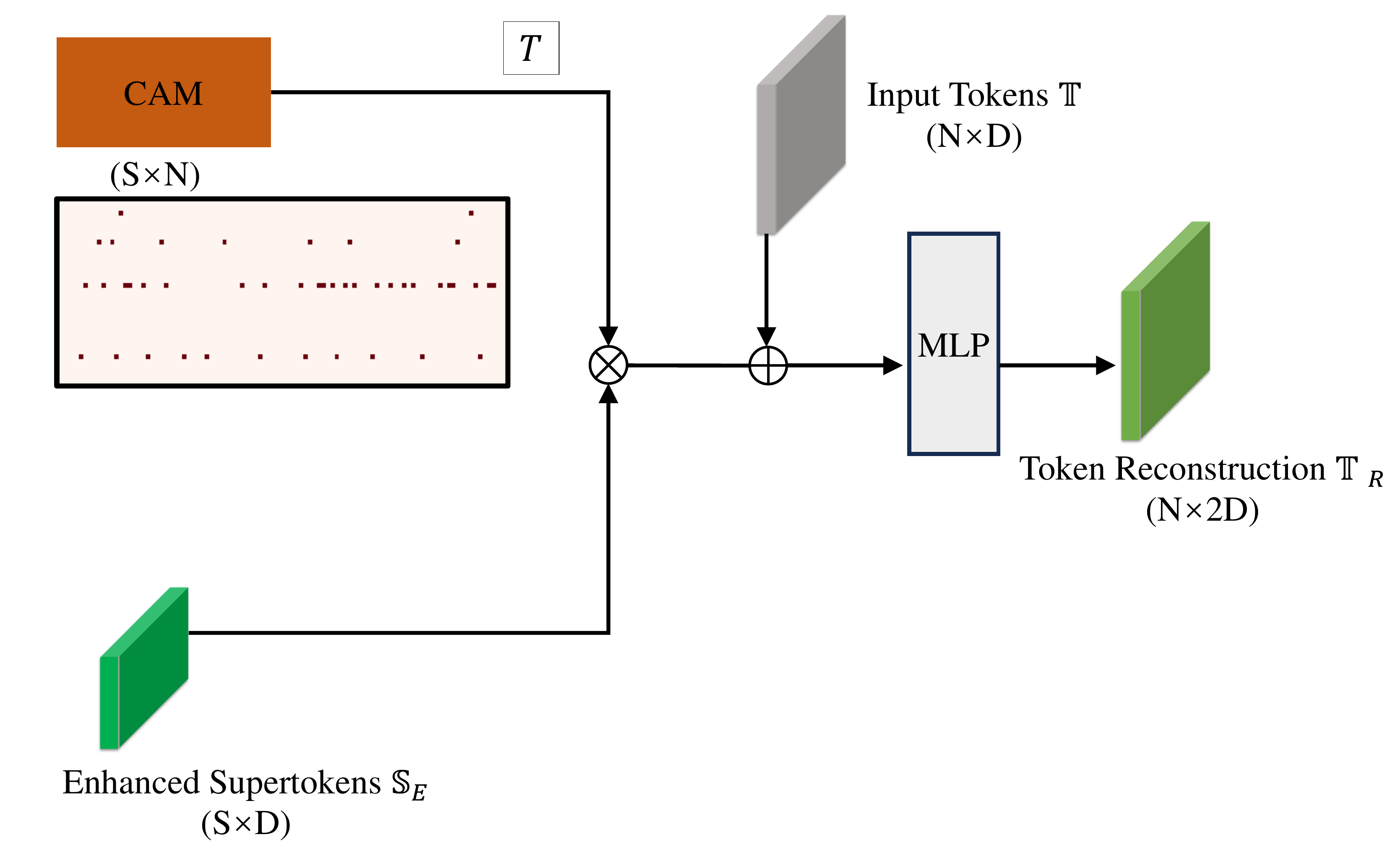}
  \caption{Illustration of the CAU block.
  \label{fig:cau}}
\end{figure}

\subsection{Deep Feature Enhancement (DFE)}
\label{subsec:DFE}
The updated $\mathbb{S}$ is further enhanced in the Transformer-based DFE block, thanks to the excellent context modeling ability of the Transformer. The design of the DFE block is shown in Fig. \ref{fig:dfe}. It fully explores the feature relationships between supertokens, which is formulated as follows:
\begin{equation}
\begin{aligned}
\label{eq:dfe}
\mathbb{S}_{E} &= \mathbb{S} + softmax(\frac{Q_{E} \times K_{E} ^{T}}{\sqrt{D}}) \times V_{E},\\
Q_{E} &= \mathbb{S} \times W_{QE},\\
K_{E} &= \mathbb{S} \times W_{KE},\\
V_{E} &= \mathbb{S} \times W_{VE},
\end{aligned}
\end{equation}
where $\mathbb{S}_{E}$ is the enhanced supertoken set, $W_{QE}$, $W_{KE}$, $W_{VE}$ are learnable weight matrices in the DFE block, which are similar to Eq. \ref{eq:qkv}. Since there are much fewer supertokens than the original input tokens, the computational complexity is reduced significantly (from $\mathcal{O}(N^{2}D)$ to $\mathcal{O}(S^{2}D)$, where $S \ll N$). Ablation experiments in Section \ref{subsec:ablation} also demonstrate the positive effect of the DFE block.

\subsection{Cross-Attention-guided Upsampling (CAU)}
\label{subsec:CAU}
\begin{figure*}[htbp]
  \centering
  \includegraphics[width=0.9\linewidth]{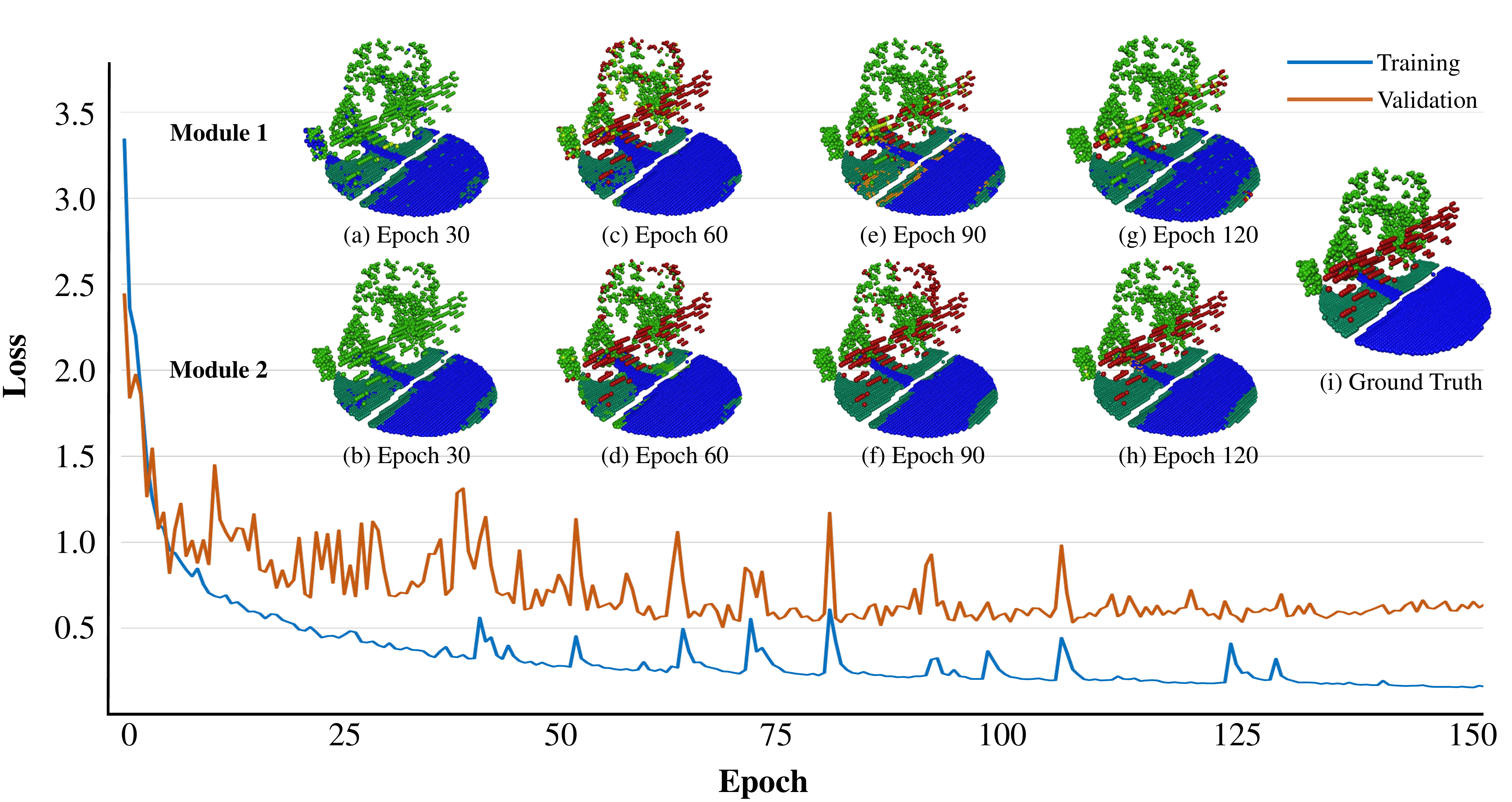}
  \caption{Token reconstruction results on each module during network training, where the training and validation loss curves are also shown. The reconstruction results gradually approach the ground truth as the network is trained, illustrating the strong feature modeling capabilities of 3DLST.
  \label{fig:loss}}
\end{figure*}

Given the enhanced supertoken set $\mathbb{S}_{E}$, we propose a simple yet effective upsampling block for token reconstruction. As shown in Fig. \ref{fig:cau}, since the clustering relationship has been recorded in the CAM, the token reconstruction process can be formulated as below.
\begin{equation}
\label{eq:cau}
\mathbb{T}_{R} = MLP(\mathbb{T} + CAM^{T} \times \mathbb{S}_{E}) \in R^{N \times 2D}.
\end{equation}
The CAU block is placed as the last block in each module of 3DLST, resulting in the novel W-net architecture. As such, the input tokens could be optimized dynamically during network training. Compared with the symmetric U-net design \citep{qi2017pointnet++}, the proposed W-net architecture allows the CAM to provide prompt guidance for token reconstruction, resulting in better prediction results. Please refer to Section \ref{subsec:ablation} for detailed comparisons and analysis. Fig .\ref{fig:loss} shows the prediction results of token reconstruction in each module. From the results, the prediction result from Module 2 is generally better than that from Module 1, which confirms the dynamic optimization of token reconstruction in 3DLST. Additionally, prediction results from both Module 1 and 2 get better and better as the network is trained, which illustrates the strong feature learning capabilities of 3DLST. 

\subsection{SuperToken Sparsification (STS)}
\label{subsec:STS}
Given the enhanced supertokens $\mathbb{S}_{E}$ of Module 1, we design a learnable supertoken sparsification block to simplify it, resulting in hierarchical feature learning for efficiency improvement. The supertokens are dropped adaptively according to the decision scores generated by their GLocal features, where GLocal features represent both global and local features of supertokens. In detail, The GLocal features of $\mathbb{S}_{E}$ are expressed as below.
\begin{equation}
\label{eq:sts}
\begin{aligned}
\mathbb{S}_{E}^{local} &=  MLP(\mathbb{S}_{E}) \in R^{S \times D}, \\
\mathbb{S}_{E}^{global} &=  Ave(MLP(\mathbb{S}_{E}) \in R^{1 \times D}, \\
\mathbb{S}_{E}^{glocal} &= \left [\mathbb{S}_{E}^{local},  Rep(\mathbb{S}_{E}^{global}) \right ] \in R^{S \times 2D}, \\
\end{aligned}
\end{equation}
where $\mathbb{S}_{E}^{local}, \mathbb{S}_{E}^{global}, \mathbb{S}_{E}^{glocal}$ represent the local, global, and GLocal embeddings of $\mathbb{S}_{E}$, respectively; $Ave$ represents the average pooling operation; $Rep$ is used to expand the size of $\mathbb{S}_{E}^{global}$ to $S \times D$ by repetition. After that, a decision score set $\Psi = \left\{ \psi _{i} \right\}_{i \in S} \in R^{S \times 2}$ is computed as:
\begin{equation}
\Psi =  Softmax(MLP(\mathbb{S}_{E}^{glocal})) \in R^{S \times 2}.
\end{equation}
We denote the two channels of $\psi _{i}$ as $\psi _{i,0}$ and $\psi _{i,0}$, which represent supertoken keeping and dropping probabilities respectively. Further, we select top-K supertokens from $\mathbb{S}_{E}^{glocal}$ with the highest keeping probabilities to form the input supertoken set $\mathbb{S}^{'}$ for Module 2 (K is set to S/2 in our experiments). To ensure a differentiable process, the Gumbel-Softmax strategy \citep{JangGP17} is employed for top-K selection. As such, the supertoken sparsification could be achieved adaptively and efficiently.

\begin{figure*}[b]
  \centering
  \includegraphics[width=0.85\linewidth]{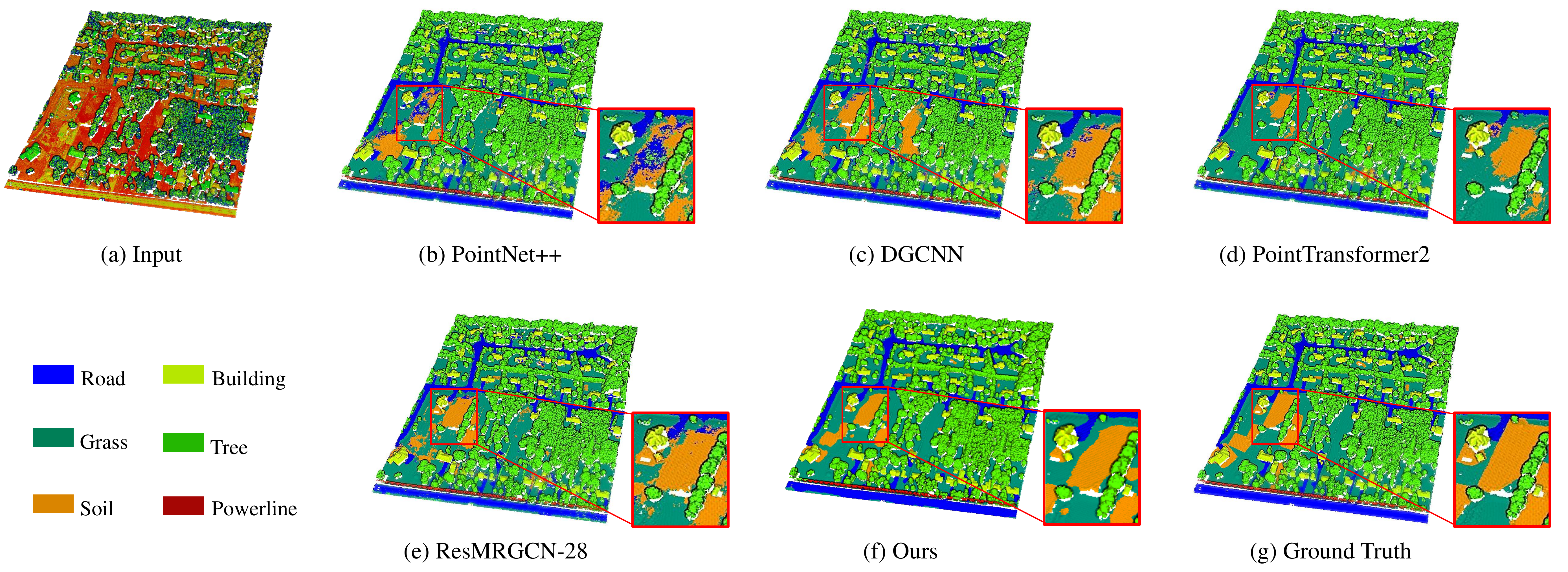}
  \caption{Comparison results from different methods on the testing area-11 in the airborne MS-LiDAR dataset. 
  \label{fig:r1}}
\end{figure*}

\begin{figure*}[htbp]
  \centering
  \includegraphics[width=0.85\linewidth]{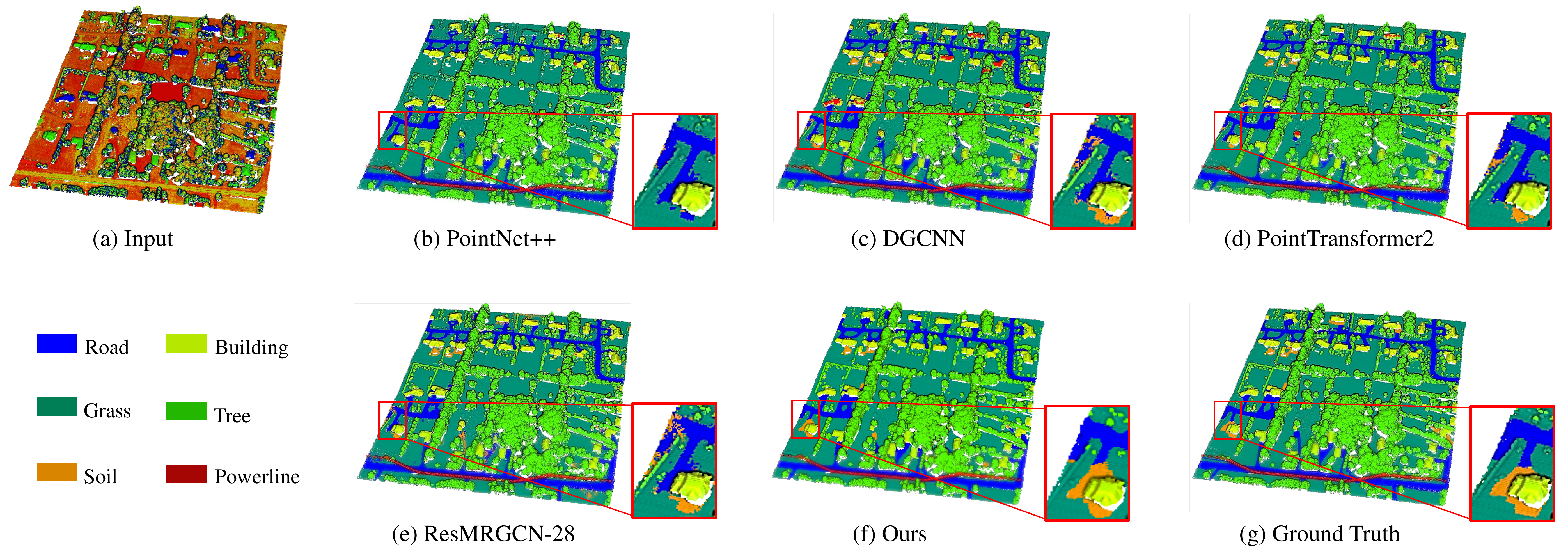}
  \caption{Comparison results from different methods on the testing area-12 in the airborne MS-LiDAR dataset.
  \label{fig:r2}}
\end{figure*}

\begin{figure*}[htbp]
  \centering
  \includegraphics[width=0.85\linewidth]{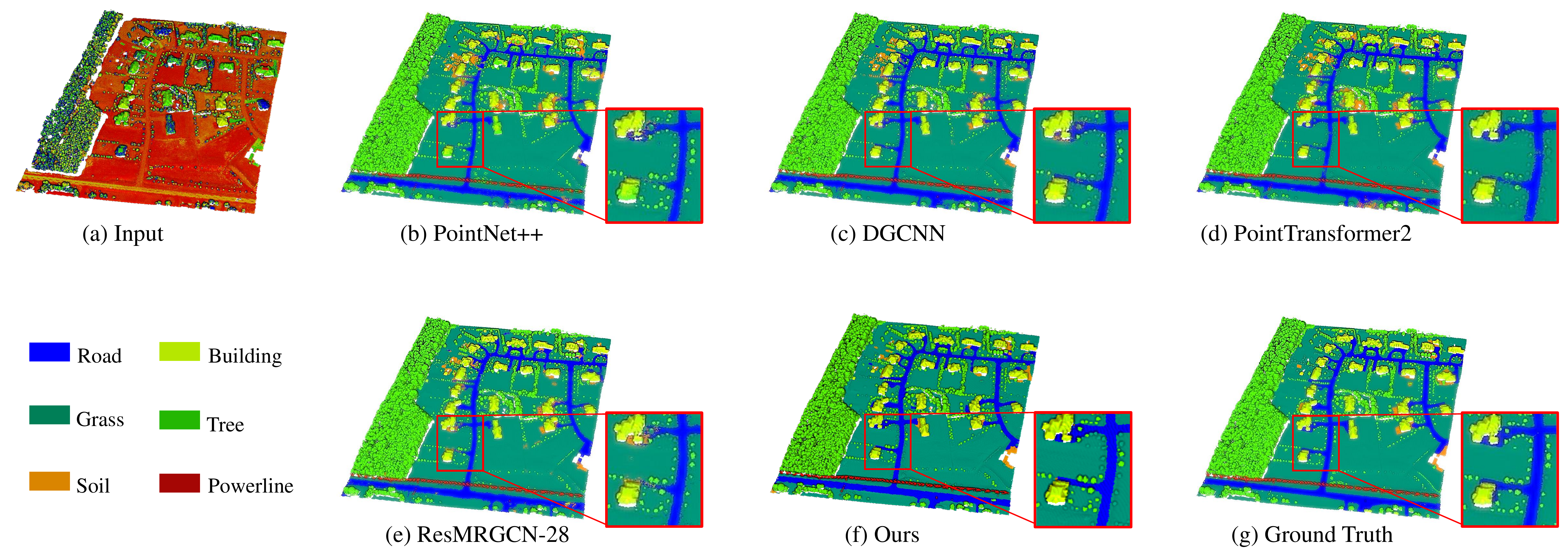}
  \caption{Comparison results from different methods on the testing area-13 in the airborne MS-LiDAR dataset.
  \label{fig:r3}}
\end{figure*}

\section{Experiments}
\label{sec:Experiments}
This section first introduces the implementation details of 3DLST, including hardware configuration and hyperparameter settings for different datasets (airborne MS-LiDAR datasets \citep{zhao2021airborne}, DALES \citep{varney2020dales}, and Toronto-3D \citep{tan2020toronto}). 
Secondly, we present the performance evaluation of our method and compare it with other state-of-the-art methods in point cloud segmentation.
Finally, we conducted ablation studies on the main components of our method, as well as explored the sensitivities of the model to parameters.

\begin{table*}[htbp]\color{black}
 \caption{Confusion matrix ($\%$) of 3DLST on the airborne MS-LiDAR dataset. The numbers from the second to seventh lines represent the number of points, while the numbers in the last three rows represent the precision, recall, and $F_{1}$ score for each class. \label{tab:ms-lidar}
 }
 \centering
 \setlength{\tabcolsep}{20pt}
 \renewcommand{\arraystretch}{1.2}
 \begin{tabular}{l|l|l|l|l|l|l}
  \hline
   {Categories}  & Road & Building & Grass  & Tree & Soil & Powerline  \\
  \hline
  {Road} & 89.0  & 2.2   & 0.0  & 0.0 &17.3 & 0.0   \\
  {Building} & 9.6  & 95.6   & 0.5  & 0.1 &28.5 & 0.0   \\
  {Grass} & 0.3  & 1.3   & 99.3  & 0.6 &0.0 & 15.1   \\
  {Tree} & 0.0  & 0.2   & 0.2  & 99.4 &0.0 & 1.0   \\
  {Soil} & 1.0  & 0.7   & 0.0  & 0.0 &54.2 & 0.0   \\
  {Powerline} & 0.0  & 0.0   & 0.0  & 0.0 &0.0 & 83.8   \\
 \hline
 {Precision} & 87.8  & 92.0   & 99.1  & 97.0 &87.2 & 96.2   \\
 {Recall} & 89.0  & 95.6  & 99.3  & 99.4 & 54.2 & 83.8   \\
 {$F_{1}$} & 88.4  & 93.8   & 99.2  & 98.2 &66.8 & 89.6   \\
  \hline
 \end{tabular}
\end{table*}

\begin{table*}[htbp]\color{black}
 \caption{Quantitative comparison ($\%$) of semantic segmentation performance on the airborne MS-LiDAR dataset. The highest evaluation score is shown in bold type. \label{tab:ms_comparison}
 }
 \centering
 \setlength{\tabcolsep}{10pt}
 \renewcommand{\arraystretch}{1.2}
 \begin{tabular}{l|l|l|l|l|l}
  \hline
   {Methods}  &Input Points & Average $F_{1}$ score & mIoU & OA & Latency ($ms$)  \\
  \hline
  {PointNet++} \citep{qi2017pointnet++}  & 4096 & 72.1 & 58.6 & 90.1   & 322.6 \\
  {DGCNN} \citep{wang2019dynamic} & 4096 & 71.6 & 51.0 & 91.4   & 86.2   \\
  {RSCNN} \citep{liu2019relation} & 4096 & 73.9 & 56.1 & 91.0  & 158.7  \\
  {GACNet} \citep{wang2019graph}  & 4096 & 67.7 & 51.0 & 90.0  & 277.8 \\
  {AGConv} \citep{zhou2021adaptive}  & 4096 & 76.9 & 71.2 & 93.3 & 312.5  \\ 
  {SE-PointNet++} \citep{jing2021multispectral}  & 4096 & 75.9 & 60.2 & 91.2  & - \\
  {FR-GCNet} \citep{zhao2021airborne} & 4096 & 78.6 & 65.8 & 93.6 & - \\
  {PointTransformer2} \citep{zhao2021point} & 4096 & 80.5 & 73.6 & 93.1 & 285.7 \\
  {PPT-Net} \citep{hui2021pyramid} & 4096 & 80.1 & 73.6 & 92.7 & 43.3 \\
  {Xiao et al.} \citep{xiao2022multispectral}  & 4096 & 83.3 & 79.3 & 94.0  & - \\
  {PatchFormer} \citep{zhang2022patchformer}  & 4096 & 82.4 & 77.8 & 93.1 & 62.9  \\
  {ResMRGCN-28} \citep{9408381}  & 4096 & 81.1 & 74.0 & 93.3  & 45.7\\
  {GCNAS} \citep{zhang2023semantic}  & 4096 & 88.1 & \textbf{82.3} & 95.2 & - \\
  {3DGTN} \citep{lu20243dgtn}  & 4096 & 88.6 & 82.1 & 95.2 & 217.4 \\
  {DCTNet} \citep{lu2024dynamic}  & 4096 & 86.0 & 80.2 & 95.0 & 53.3 \\
  \hline
  {Ours}   & 4096  &\textbf{89.3} & \textbf{82.3 }& \textbf{95.4} & \textbf{11.7}\\

  \hline
 \end{tabular}
\end{table*}

\begin{figure*}[htbp]
  \centering
  \includegraphics[width=0.9\linewidth]{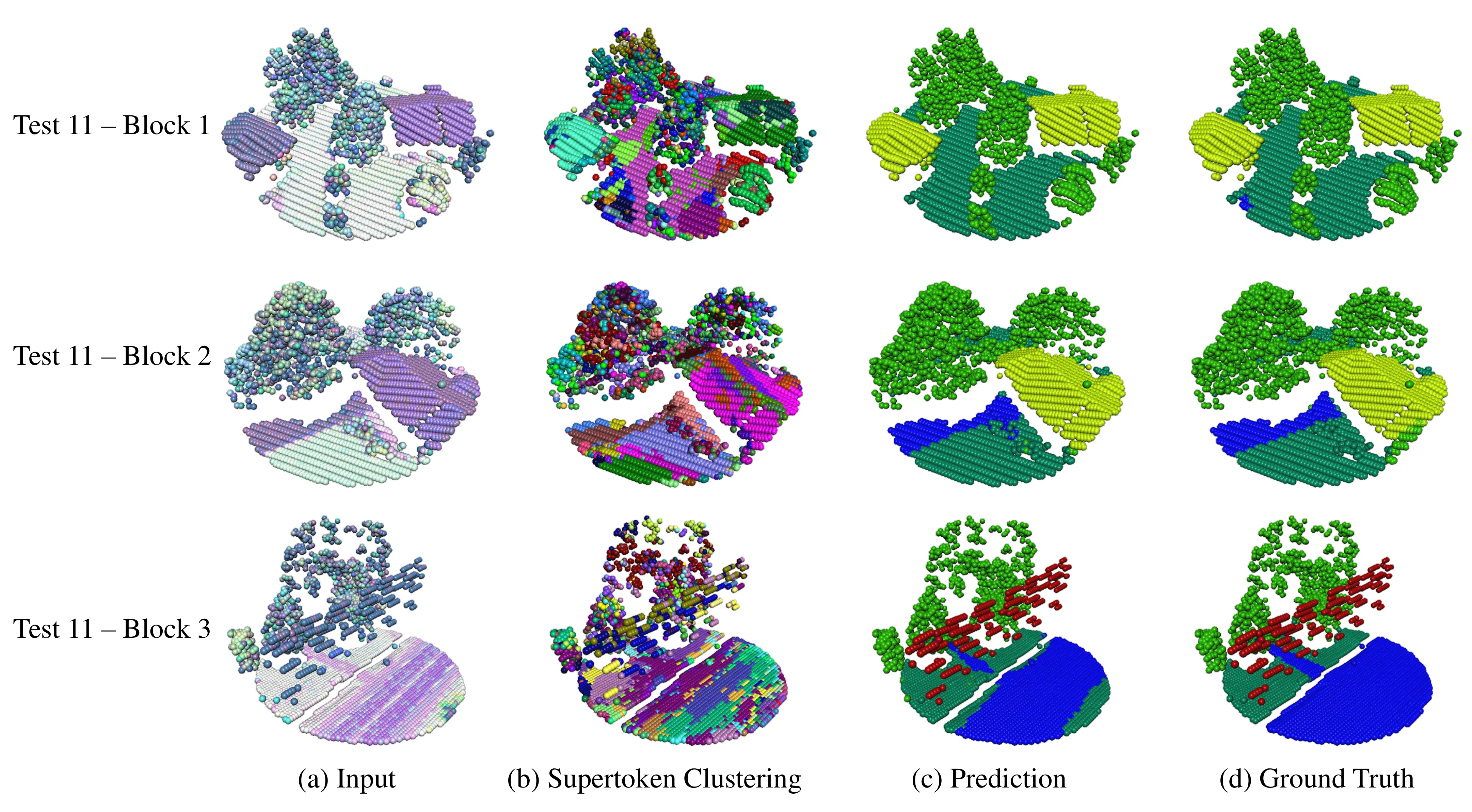}
  \caption{Visualization results of supertoken clustering and prediction on the testing area-11 blocks. They confirm the excellent performance of 3DLST in semantic-homogeneity token clustering and prediction on the airborne MS-LiDAR data.
  \label{fig:overseg}}
\end{figure*}

\subsection{Implementation Details}
3DLST was implemented in PyTorch and trained/tested on an NVIDIA Tesla V100 GPU. It was trained with the SGD Optimizer, with a momentum of 0.9 and weight decay of 0.0001. The initial learning rate was set to 0.01, with a cosine annealing schedule for learning rate adjusting. The batch size was set to 16 for the airborne MS-LiDAR and Toronto-3D datasets, and 8 for the DALES datasets, according to different input sizes. For each dataset, the network was trained for 250 epochs, with the weighted Cross-Entropy (CE) loss function.

\begin{table}[htbp]
 \caption{Quantitative comparison ($\%$) of semantic segmentation performance on the DALES dataset, including OA and mIoU. The highest evaluation score is shown in bold type. \label{tab:dales}
 }
 \centering
 \setlength{\tabcolsep}{6pt}
  \renewcommand{\arraystretch}{1.2}
 \begin{tabular}{l|l|l}
  \hline
   {Methods} & OA & mIoU \\
  \hline
  {PointNet++}\citep{qi2017pointnet++} & 95.7  & 68.3 \\
  {ConvPoint}\citep{boulch2020convpoint} & 97.2 & 67.4 \\
  {SPG}\citep{landrieu2018large} & 95.5 &60.6 \\
  {PointCNN}\citep{li2018pointcnn} & 97.2 &58.4 \\
  {ShellNet}\citep{zhang2019shellnet} & 96.4 &57.4    \\
  {SPT}\citep{robert2023efficient} & 97.5 &	79.6  \\
  {SuperCluster}\citep{robert2024scalable} & - &77.3\\
  \hline
  {Ours}    & \textbf{97.6}  & \textbf{80.2 }\\

  \hline
 \end{tabular}
\end{table}

\begin{table*}[htbp]\color{black}
 \caption{Detailed comparison between SPT \citep{robert2023efficient} and 3DLST on the DALES dataset, including OA and mIoU. In addition, the preprocessing time and inference time for the testing set are also reported.\label{tab:dales_spt}
 }
 \centering
 \setlength{\tabcolsep}{15pt}
  \renewcommand{\arraystretch}{1.2}
 \begin{tabular}{l|l|l|l|l}
  \hline
   {Methods} & OA & mIoU & Preprocessing (s) & Inference (s) \\
  \hline
  {SPT}\citep{robert2023efficient} & 97.5 &	79.6 & 125.3 & \textbf{2.0}  \\

  \hline
  {\textbf{3DLST}}    & \textbf{97.6}  & \textbf{80.2}  & \textbf{76.1} & 2.36 \\

  \hline
 \end{tabular}
\end{table*}

\begin{figure*}[htbp]
  \centering
  \includegraphics[width=0.9\linewidth]{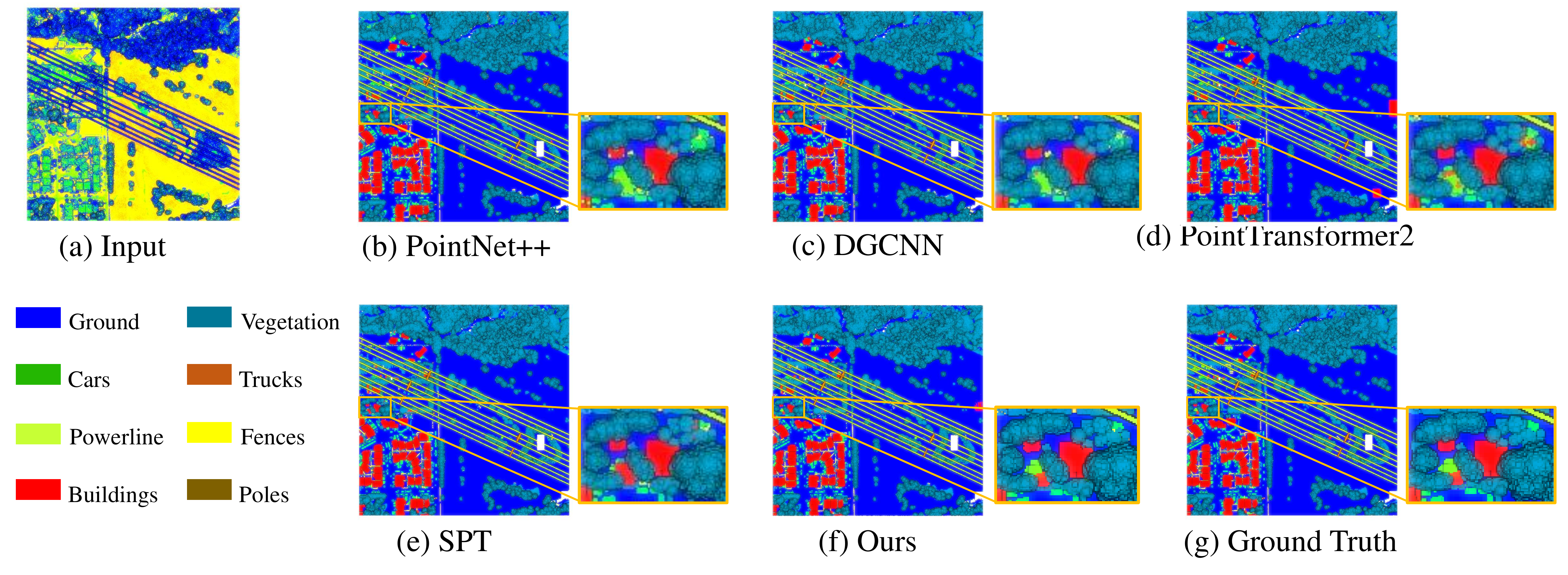}
  \caption{Comparison results from different methods on the DALES dataset. As can be seen, our segmentation predictions are faithful to the ground truth.
  \label{fig:dales}}
\end{figure*}

\begin{table*}[htbp]\color{black}
 \caption{Quantitative comparison ($\%$) of semantic segmentation performance on the Toronto-3D dataset, including OA, mIoU, and the IoU value for each category. The highest evaluation score is shown in bold type. \label{tab:toronto3d}
 }
 \centering
 \setlength{\tabcolsep}{4pt}
  \renewcommand{\arraystretch}{1.2}
 \begin{tabular}{l|l|l|l|l|l|l|l|l|l|l}
  \hline
   {Methods} & OA & mIoU & Road & Rd mrk. & Natural & Building & Util. line & Pole & Car & Fence \\
  \hline
  {PointNet}\citep{qi2017pointnet} & 92.6  & 59.5 & 92.9 &0.0 &86.1 &82.1 &60.9 &62.8 &76.4 &14.4   \\
  {DGCNN}\citep{wang2019dynamic} & 94.2 &61.8 &93.8 &0.0 &91.2 &80.3 &62.4 &62.3 &88.2 &15.8   \\
  {MS-PCNN}\citep{ma2019multi} & 90.0 &65.9 &93.8 &3.8 &93.4 &82.5 &67.8 &71.9 &91.1 &22.5   \\
  {TGNet}\citep{li2019tgnet} & 94.0 &61.3 &93.5 &0.0 &90.8 &81.5 &65.2 &62.9 &88.7 &7.8   \\
  {KPConv}\citep{thomas2019kpconv} & 95.3 &69.1 &94.6 &0.0 &96.0 &91.5 &\textbf{87.6} &81.5 &85.6 &15.7    \\
  {MS-TGNet}\citep{tan2020toronto} & 95.7 &70.5 &94.4 &17.1 &95.7 &88.8 &76.0 &73.9 &\textbf{94.2} &23.6   \\
  {AF-GCN}\citep{zhang2023improving} & 97.0 &79.7 &\textbf{97.4} &\textbf{69.5} &94.7 &\textbf{94.9} &78.2 &\textbf{83.3} &91.5 &28.2  \\
  \hline
  {Ours}    & \textbf{97.2}  & \textbf{80.4 } &96.9 & 64.3 & \textbf{96.1} & 91.1 & 79.1 &71.6 & 92.8 & \textbf{51.2 }    \\

  \hline
 \end{tabular}
\end{table*}

\subsection{Airborne MS-LiDAR Dataset}
\textbf{Dataset and Metrics.} 
The large-scale airborne MS-LiDAR dataset was introduced in \cite{zhao2021airborne}. The dataset was collected by a Teledyne Optech Titan MS-LiDAR system, containing 13 areas in total. Specifically, the training set contains No.1-10 areas, and the testing set contains No.11-13 areas. Each area covers over 15000 $m^{2}$ with an average point density of about 3.6 points/$m^{2}$. Each point has six attributes: XYZ, MIR (1550 $nm$ wavelength), NIR (1064 $nm$ wavelength), and Green (532 $nm$ wavelength). There are six categories in the datasets: Road, Building, Grass, Tree, Soil, and Powerline. The imbalanced class distribution makes the dataset more challenging, where the number of road points is more than 30$\times$ the number of powerline points. Similar data pre-processing methods to \cite{zhao2021airborne} are applied in our experiments, ensuring a fair comparison. Each area was split into a series of local blocks by $k$-nearest neighbor ($k$nn) searching. $k$ is set to 4096 in our experiments. The mean Intersection over Union (mIoU), Overall Accuracy (OA), average $F_{1}$ score, and latency are used for performance evaluation, where the latency refers to the time ($ms$) required for a model to process the input data. Besides, we also provided the Precision, Recall, and $F_{1}$ score for each category. For a fair comparison, the indicator of latency is measured on an NVIDIA GeForce RTX 2080 GPU, and the batch size of the input points is set to 1.

\textbf{Performance Comparison.} 
A detailed performance evaluation of 3DLST on the airborne MS-LiDAR dataset is shown in Table \ref{tab:ms-lidar}, presented as a confusion matrix. The Precision, Recall, and the $F_{1}$ score for each category are presented in the last three rows of the table. From the results, 3DLST achieves the $F_{1}$ scores of over 85$\%$ for all categories except soil, where the $F_{1}$ scores of grass and tree exceeds 98$\%$. Moreover, comparison results shown in Table \ref{tab:ms_comparison} demonstrate that 3DLST surpasses the previous SOTA methods in terms of both accuracy and efficiency. Specifically, 3DLST is over 5$\times$ faster than PatchFormer \citep{zhang2022patchformer}, and about 4$\times$ faster than ResMRGCN-28 \citep{9408381}. The visualization of the segmentation result comparison is shown in Fig. \ref{fig:r1}, \ref{fig:r2}, and \ref{fig:r3}, showing the excellent segmentation performance of 3DLST. Additionally, we also show the clustering results of supertokens in 3DLST in Fig.\ref{fig:overseg}. Compared with the traditional superpoint clustering \citep{landrieu2018large}, our clustering results are visually scattered. This is because the proposed supertokens focus on splitting the input samples into a series of semantic-homogeneity groups, instead of geometry-based clustering. Overall, both comparison and visualization results demonstrate the superiority of 3DLST in airborne MS-LiDAR point cloud scene segmentation.

\subsection{DALES Dataset}
\textbf{Dataset and Metrics.} 
Dayton Annotated LiDAR Earth Scan (DALES) dataset was presented by \cite{varney2020dales}. It is a new large-scale aerial LiDAR dataset with over a half-billion points spread across 10 square kilometers, which can be categorized into eight distinct object classes: Ground, Vegetation, Cars, Trucks, Powerlines, Fences, Poles, and Buildings. The data was collected using a Riegl Q1560 dual-channel system flown in a Piper PA31 Panther Navajo. Point cloud data obtained from Aerial Laser Scanners (ALS) present unique challenges and offer novel applications, particularly in fields like 3D urban modeling and large-scale surveillance. The DALES dataset was split into 40 areas with the elaborating class annotation, each spanning 0.5 $km^{2}$ and containing 12 million points. Each input point has 6 attributes (XYZ + RGB). For a fair comparison, we subsample each area using a 10 $cm$ grid and split the 40 areas into training and testing sets with roughly a 70/30 percentage. For each area, we divided it into a series of $20 m \times 20 m$ blocks as training/testing samples, where each of them contained 8192 points after sampling. The mIoU and OA were used for performance evaluation.

\textbf{Performance Comparison.} 
Comparison results of different methods on the DALES dataset are shown in Table \ref{tab:dales}. We compared 3DLST with other SOTA point cloud segmentation methods including the most recent superpoint Transformer method, SPT \citep{robert2023efficient}. From the results, our method outperforms other benchmarked SOTA methods, with the highest mIoU (80.2$\%$) and OA (97.6$\%$). This demonstrates the excellent segmentation performance of 3DLST to aerial LiDAR data. Moreover, we also compared the inference speed of 3DLST and SPT \citep{robert2023efficient} on one testing area ($5135\_54435$) in detail. As shown in Table \ref{tab:dales_spt}, 3DLST is over 2$\times$ faster than SPT in terms of the preprocessing step. This is because 3DLST does not require the time-consuming superpoint generation process. Since the size of input tokens to SPT is smaller than ours (by the preprocessing step of superpoint clustering), the network inference speed of 3DLST is slightly lower than SPT. Overall, the entire inference speed of 3DLST is 1.6$\times$ faster than SPT, while achieving a slightly higher mIoU. This demonstrates that 3DLST achieves SOTA performance in both algorithm accuracy and speed. The visualization of the segmentation result comparison on the DALES dataset is shown in Fig. \ref{fig:dales}, which confirms the superiority of 3DLST in aerial LiDAR point cloud segmentation.

\subsection{Toronto-3D Dataset}
\textbf{Datasets and Metrics.} 
Toronto-3D dataset was collected by a vehicle-mounted MLS system (Teledyne Optech Maverick) in large-scale urban outdoor scenarios. It consists of more than $78$ million points, covering approximately 1 $km$ of road. There are 8 categories included in the dataset: Road, Road marking, Natural, Building, Utility line, Pole, Car, and Fence. For a fair comparison, we split the dataset into four subsets: $L001, L002, L003, L004$, where $L002$ was used for testing. We subsample the datasets using a 6 $cm$ grid, followed by applying a coordinate offset as described in \cite{tan2020toronto}. Further, each subset was divided into a series of $5 m \times 5 m$ blocks and each of them contained 4096 points after sampling. Each input point has 7 attributes: XYZ, RGB, and intensity. To be consistent with the evaluation metrics used by SOTA methods, we used mIoU and OA for performance evaluation. We also provide the IoU value for each category. 

\textbf{Performance Comparison.} 
The comparison results are shown in Table. \ref{tab:toronto3d}. From the results, 3DLST achieved the highest mIoU (80.4$\%$) compared with other SOTA methods such as AF-GCN \citep{zhang2023improving}, as well as competitive OA (97.2$\%$). Specifically, 3DLST achieved over an 80$\%$ improvement in IoU on the $Fence$ class. These results demonstrate the excellent performance of 3DLST on vehicle-mounted LiDAR data processing.

\subsection{Ablation Study}
\label{subsec:ablation}
A series of ablation studies were conducted for the main blocks and parameters of 3DLST, to verify the effectiveness of the proposed blocks and robustness of the network to parameters. These experiments were performed on the airborne MS-LiDAR dataset.

\begin{table*}[htbp]\color{black}
 \centering
 \setlength{\tabcolsep}{6pt}
  \renewcommand{\arraystretch}{1.2} 
 \caption{\textcolor{black}{Results ($\%$) of ablation study on the airborne MS-Lidar dataset} 
 }
 \label{tab:ablation}
 \begin{tabular}{l|l|l|l|l|l}
  \hline
    \multicolumn{2}{c|}{Ablation}  & Average $F_{1}$ score & mIoU & OA & Latency ($ms$)\\
 \hline

  {Dynamic Supertoken Optimization} & $argmax$ $\rightarrow$ $softmax$  & 87.1  & 79.9 & 94.6 & 11.5     \\
   \hline
  {Deep Feature Enhancement} & $-$  & 82.5  & 71.7 & 91.6 & \textbf{9.8  }   \\
  \hline
   {Network Architecture} & $W$-net design $\rightarrow$$ U$-net design  & 82.6  & 70.9 & 91.2 & 11.3     \\
   \hline
   \hline

  \multicolumn{2}{c|}{3DLST}  &\textbf{89.3} & \textbf{82.3 }& \textbf{95.4} & 11.7\\
  \hline
 \end{tabular}
\end{table*}

\begin{table}[htbp]
 \caption{Parameter sensitivity of 3DLST to the number $S$ of learnable supertokens on the airborne MS-LiDAR dataset. \label{tab:number_s}
 }
 \centering
 \setlength{\tabcolsep}{8pt}
 \renewcommand{\arraystretch}{1.2}
 \begin{tabular}{l|l|l|l}
  \hline
   {$S$} & Average $F_{1}$ score ($\%$) & mIoU ($\%$) & OA ($\%$)  \\
  \hline
  128  & 80.1 & 69.6 & 93.1    \\
  256  & 87.8 & 80.3 & 94.5   \\
  \textbf{512}  & \textbf{89.3} & \textbf{82.3} & \textbf{95.4}  \\
  1024  & 86.4 & 78.1 & 93.9    \\
  \hline
 \end{tabular}
\end{table}

\textbf{Dynamic Supertoken Optimization.} 
As described in Section \ref{subsec:DSO}, we proposed a cross-attention mechanism with a hard assignment for learnable supertoken optimization. Firstly, in the process of CAM generation, we replaced the $argmax$ operator in Eq. \ref{eq:cam} with $softmax$. As shown in Table \ref{tab:ablation} Row 2, average $F_{1}$ score, mIoU, and OA values dropped 2.2, 2.4, and 0.8 absolute percentage points respectively after replacing. Compared with $softmax$, $argmax$ focuses on discrete decision-making, which is more suitable for our supertoken clustering.

Secondly, we tested the sensitivity of networks to the number $S$ of learnable supertokens. As shown in Table \ref{tab:number_s}, the best performance of 3DLST on the airborne MS-LiDAR dataset is achieved when $S$ is set to 512. On the one hand, the smaller $S$ may cause the network to ignore detailed features in a scene, resulting in inaccurate boundary segmentation and category misclassification. On the other hand, increasing $S$ tends to limit the ability of the network to capture sufficient contextual information.

\textbf{Deep Feature Enhancement.} 
We removed the DFE block from 3DLST to verify its effectiveness. As shown in Table \ref{tab:ablation} Row 3, the mIoU of 3DLST after removing the DFE block dropped from 82.3$\%$ to 71.7$\%$, which demonstrates the positive effect of the DFE block in feature enhancement. Accordingly, the latency of the network was reduced slightly, from 11.7 $ms$ to 9.8 $ms$.



\textbf{W-net Architecture.}
Table \ref{tab:ablation} Row 6 shows the performance comparison of W-net and U-net \citep{qi2017pointnet++} architectures in 3DLST. From the results, the performance of the U-net drops significantly compared with our W-net. This is because the CAM-based token reconstruction in the U-net is lagging and inaccurate. It is challenging for the cross-attention map in the shallow module to accurately describe the feature relationships in the decoder. In contrast, the proposed W-net architecture allows the generated CAM to be promptly fed back to the supertokens in the same module for token reconstruction, which ensures the timeliness of feature interaction and reconstruction.

\section{Conclusion}
\label{sec:conclusion}
In this work, we proposed a novel 3D Transformer method for LiDAR point cloud scene segmentation, named 3DLST. The main novelties of 3DLST are summarized as follows. Firstly, a novel DSO block is proposed for efficient token clustering and aggregating. Compared with the static superpoint strategy operating on the initial point features, the DSO block focuses on semantic homogeneity-aware token clustering, allowing the supertokens to be optimized dynamically according to multi-level deep features. Additionally, the DSO block avoids the time-consuming process of traditional superpoint generation, thanks to the learnable supertoken definition. Therefore, the pre-processing efficiency of 3DLST is over  2$\times$ faster than other SOTA superpoint Transformers \citep{robert2023efficient}. Secondly, we design an efficient CAU block that fully exploits the long-range context dependency modeling capability of the cross-attention mechanism for efficient and accurate token reconstruction. Ablation studies demonstrate that the CAU block achieves better results in 3DLST than traditional upsampling approaches such as Trilinear and nearest neighbor interpolation. Thirdly, based on the aforementioned blocks, we design a novel W-net architecture for 3DLST. Compared with the common U-net design, the W-net is more suitable for Transformer-based feature learning. The related ablation experiments also confirmed this. The 3DLST equipped with W-net outperforms that with U-net by 11.4 absolute percentage points in terms of mIoU. Extensive comparison experiments on three challenging LiDAR datasets (airborne MS-LiDAR, DALES, and Toronto-3D datasets) demonstrate the superiority of 3DLST over previous SOTA methods in terms of both algorithm efficiency and accuracy, as well as its strong adaptability to different types of LiDAR point cloud data (airborne MS-LiDAR, aerial LiDAR, and vehicle-mounted LiDAR data)




\bibliographystyle{cas-model2-names}
\bibliography{mybibfile}

\end{document}